\documentclass{article}
\usepackage{spconf,amsmath,epsfig}
\usepackage{cite}
\usepackage{amsmath,amssymb,amsfonts}
\usepackage{algorithmic}
\usepackage{graphicx}
\usepackage{textcomp}
\usepackage{xcolor}
\usepackage{booktabs}
\usepackage{multirow}
\usepackage{amssymb}
\usepackage{caption}
\usepackage{bbding}
\usepackage{cleveref}
\usepackage{subcaption}

\let\OLDthebibliography\thebibliography
\renewcommand\thebibliography[1]{
 \OLDthebibliography{#1}
 \setlength{\parskip}{0pt}
 \setlength{\itemsep}{0pt plus 0.3ex}
}

\begin{document}\sloppy

\def\x{{\mathbf x}}
\def\L{{\cal L}}

\title{rFaceNet: An End-to-End Network for Enhanced Physiological Signal Extraction through Identity-Specific Facial Contours}
%

\name{
Dali Zhu, Wenli Zhang, Hualin Zeng, Xiaohao Liu, Long Yang, Jiaqi Zheng}
\address{Institute of Information Engineering, Chinese Academy of Sciences, Beijing, China\\
School of Cyber Security, University of Chinese Academy of Sciences, Beijing, China\\
\{zhudali, zhangwenli, zenghualin, liuxiaohao, yanglong, zhengjiaqi\}@iie.ac.cn 
}


\maketitle
\vspace{-1cm} 

\begin{abstract}
Remote photoplethysmography (rPPG)  technique extracts blood volume pulse (BVP) signals from subtle pixel changes in video frames. This study introduces rFaceNet, an advanced rPPG method that enhances the extraction of facial BVP signals with a focus on facial contours.  rFaceNet integrates identity-specific facial contour information and eliminates redundant data. It efficiently extracts facial contours from temporally normalized frame inputs through a Temporal Compressor Unit (TCU) and steers the model focus to relevant facial regions by using the Cross-Task Feature Combiner (CTFC). Through elaborate training, the quality and interpretability of facial physiological signals extracted by rFaceNet are greatly improved compared to previous methods. Moreover, our novel approach demonstrates superior performance than SOTA methods in various heart rate estimation benchmarks.
\end{abstract}
\section{Introduction}
\label{sec:intro}

The traditional method of extracting heart rate signals via electrocardiogram (ECG) is gradually being replaced by portable, non-contact photoplethysmographic blood pressure measurements. 
Wherein, remote photovolumetric pulse (rPPG)~\cite{verkruysse2008remote} is a technique for non-invasive acquisition of blood volume pulse (BVP) signals based on skin reflection modeling, which can be further used for the extraction of heart rate and respiration signals. 
Although mainly using facial video as input, 
previous solutions, which adopt traditional signal processing techniques~\cite{tulyakov2016self,wang2016algorithmic,poh2010advancements, de2013robust,pilz2018local,de2014improved} with heuristic rules and non-end-to-end deep learning~\cite{niu2019rhythmnet, niu2019robust, lu2021hr, niu2020video, lu2021dual} with complex preprocessing, have failed to integrate with facial contours due to technical limitations. 
This integration has recently been shown an effectively enhancement for facial BVP signal extraction~\cite{chen2018deepphys,liu2020multi,liu2023efficientphys}. 

\begin{figure}[t]
\centering
\footnotesize
 \includegraphics[page=1,width=0.5\textwidth]{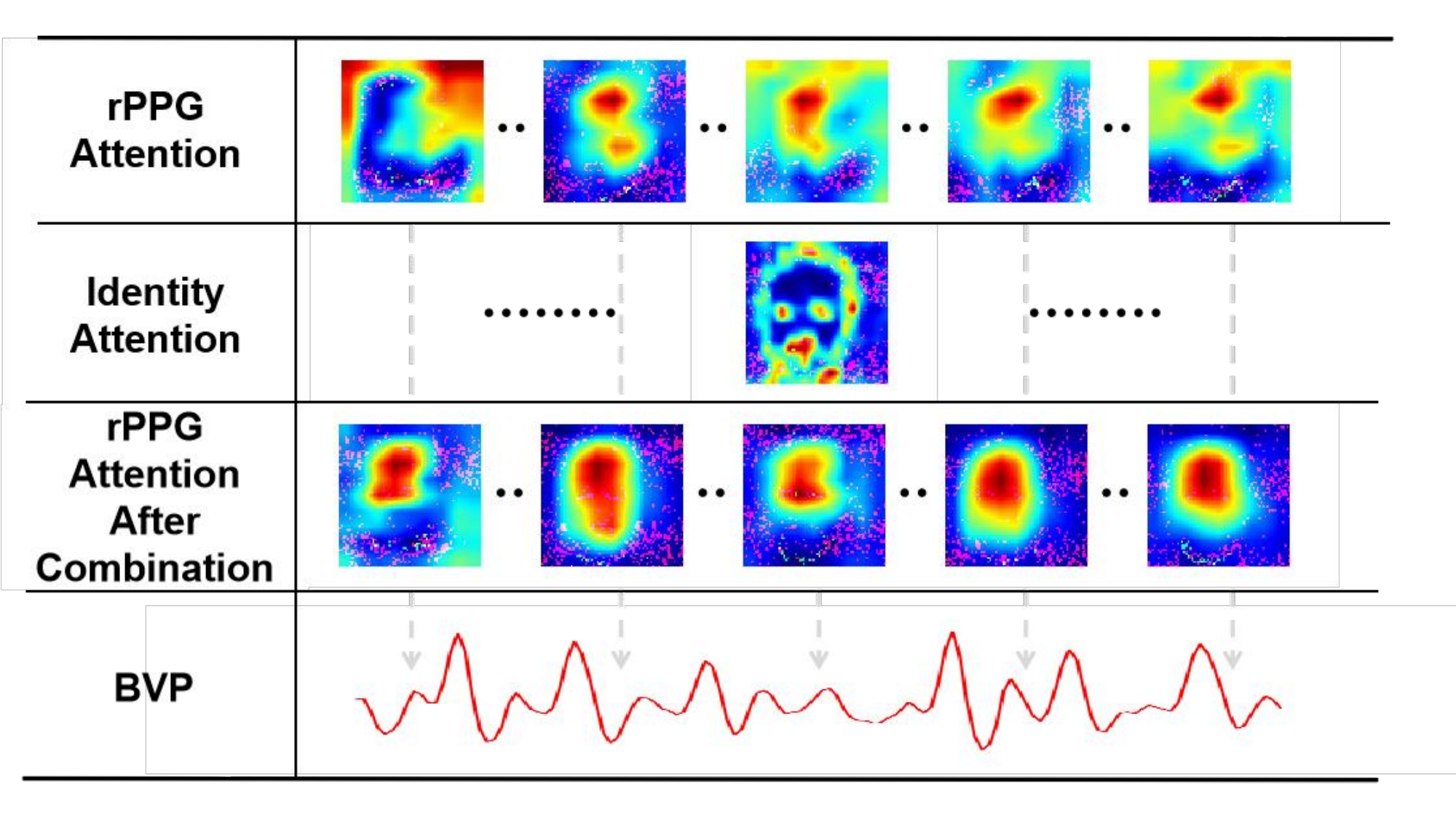}
 \caption{The rPPG attention region before and after being combined with identity-attention facial contours is shown with Gram-CAM~\cite{selvaraju2017grad}.}
 \label{fig:gram-cam}
 \vspace{-0.4cm} 
\end{figure}

Despite the significant results, 
previous works~\cite{chen2018deepphys,liu2020multi,liu2023efficientphys} tend to introduce unnecessary noises of input information when adding an equivalent number of RGB image frames, leading to the efficiency dropping. 
To minimize the redundancy, we attempt to categorize facial contours based on identity within the slight facial changes from frame to frame.
Inspired by the decoupling strategy of identity features from the inputs~\cite{chung2022domain}, we develop an innovative end-to-end framework which strategically extracts identity-specific facial contours from temporally normalized frames, thereby reducing storage space and mitigating the negative impact of noise. Additionally, we design a novel fusion approach to integrate facial contour point information into physiological signal feature maps along the temporal dimension, enabling the model to gradually learn the relationship between facial contours and physiological signals. 
As shown in Fig~\ref{fig:gram-cam}, this approach steers the model's focus to the facial contour region, ultimately improving its ability and interpretability to extract accurate physiological signals. 
Moreover, we employ an enhanced maximum magnitude multi-task loss to maintain a balance between the extraction of facial contours and BVP signals during the training. 
Extensive experiments are conducted on multiple benchmark datasets, demonstrating the rationality of our design and the superior performance than state-of-the-art methods in the heart rate estimation task.

To sum up, the main contributions of our work are threefold as follows:
\begin{itemize}
\item We propose rFaceNet for physiological signal extraction. To the best of our knowledge, we are among the first to simultaneously extract facial contours and facial physiological signals, avoiding redundant inputs and unnecessary noise.

\item We propose an innovative mechanism to integrate facial contour point information into physiological signal feature maps along the temporal dimension, which allows the model to focus on the facial contour and its interior.

\item We use an enhanced maximum magnitude multi-task loss. After elaborate training, rFaceNet outperforms SOTA methods in several benchmark datasets for the heart rate estimation task.
\end{itemize}

\section{Related Work}
\label{sec:related work}
In 2008, Verkruysse et al.~\cite{verkruysse2008remote} proposed a method for heart rate measurement using a home camera. Based on this research, more and more heart rate measurement methods based on rPPG technology were proposed. The traditional signal processing scheme~\cite{tulyakov2016self,wang2016algorithmic,de2013robust,poh2010advancements,pilz2018local,de2014improved} learned signals of different color channels through heuristics to explore the RGB images and reduced the effect of noise using techniques such as automatic face tracking and blind source separation. rPPG techniques were proven to efficiently restore the heart rate from video inputs. 
Non-end-to-end deep learning solutions were mainly used to improve the model learning efficiency by constructing spatio-temporal feature maps~\cite{niu2019rhythmnet,niu2019robust,lu2021hr,niu2020video,lu2021dual}. Meanwhile, end-to-end deep learning models were introduced into the rPPG domain~\cite{yu2023physformer++,yu2022physformer,carreira2017quo,yu2019remote1,chen2018deepphys,yu2020autohr,liu2020multi,liu2023efficientphys}. Existing techniquesms were sufficiently sophisticated in extracting heart rate estimates. However, increasingly complex preprocessing schemes and models introduced more non-interpretability and imposed more stringent requirements on the dataset. Some recent work~\cite{chen2018deepphys,liu2020multi,liu2023efficientphys} began to recognize the importance of facial appearance cues and enhanced the effectiveness and interpretability of the models by incorporating additional appearance inputs. However, attempts to extract facial contours from uniform temporal difference inputs have not emerged.

\section{Methodology}
\label{sec:methodology}
\begin{figure*}[t]
 \centering
 \footnotesize
 \includegraphics[page=2, width=1\textwidth, trim=2cm 3cm 2cm 2cm, clip]{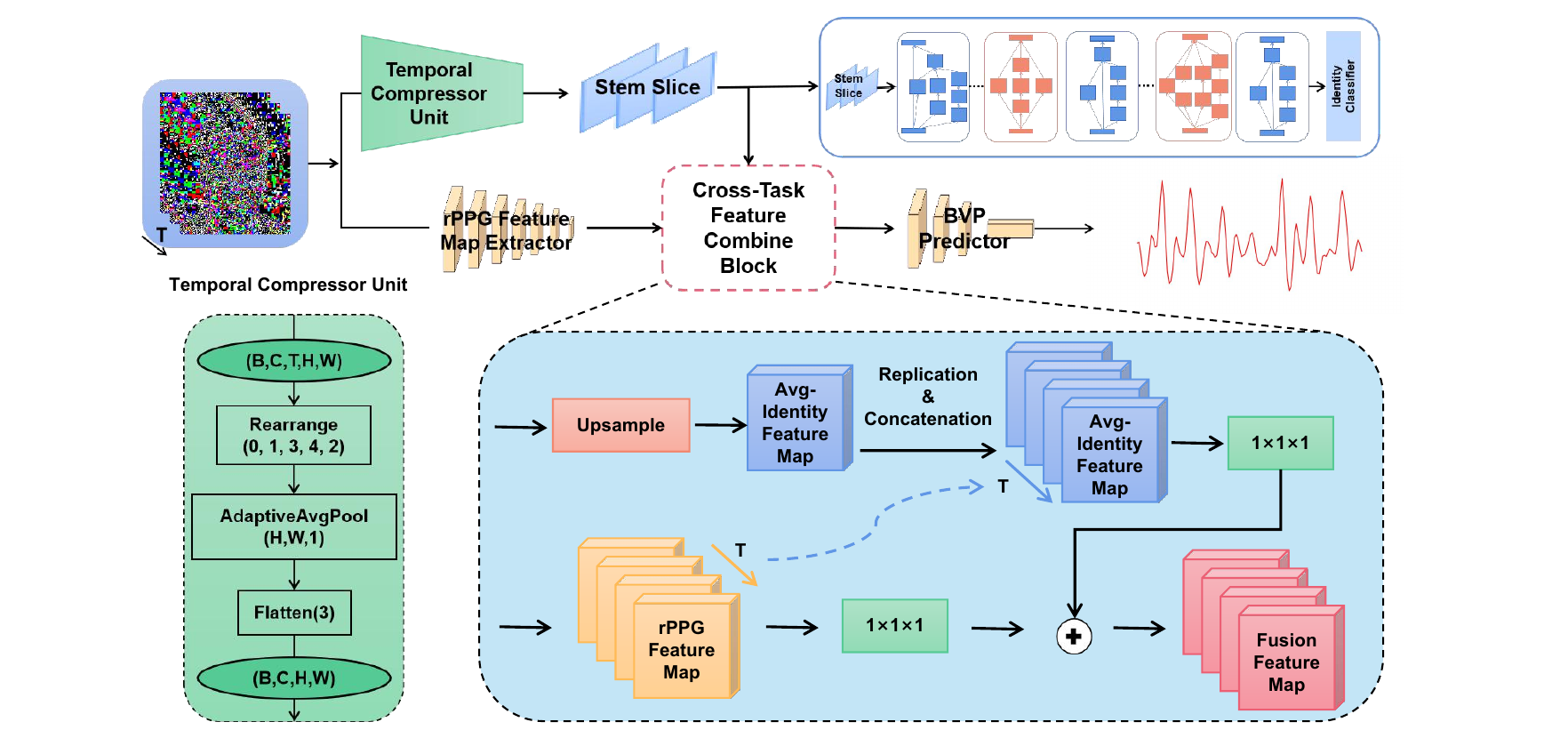}
 \caption{The framework of rFaceNet, which uses temporally normalized frames as inputs. The Temporal Compressor Unit and the Cross-Task Feature Combiner are shown in detail.}
 \label{fig:framework}
  \vspace{-0.4cm}
\end{figure*}

In this section, we begin with an overall introduction to the rFaceNet framework in Sec~\ref{sec:rFaceNet}. Then we provide a more detailed description of the Temporal Compressor Unit (TCU) in Sec~\ref{sec:temporal compressor unit}, Cross-Task Feature Combiner (CTFC) in Sec~\ref{sec:cross-task feature combine block}. Finally, we show the multi-task loss function in Sec~\ref{sec:multi-task loss}.

\subsection{rFaceNet}
\label{sec:rFaceNet}
Fig~\ref{fig:framework} shows an illustration of our rFaceNet, which is composed of two distinct branches. One branch is used to extract facial contours from different identities and includes a temporal compression unit (Sec~\ref{sec:temporal compressor unit}), a 2D feature extractor and an identity classifier. Note that the extraction of facial contours is implemented for different identities by means of the identity classifier. The other branch predicts BVP signals and includes a Cross-Task Feature Combiner(Sec~\ref{sec:cross-task feature combine block}), a 3D feature extractor and a BVP predictor.

\subsection{Temporal Compressor Unit (TCU)}
\label{sec:temporal compressor unit}
The Temporal Compressor Unit first shifts the dimensions of frames to move the spatio-temporal dimension to the end. Next, it compresses temporal dimension using 3D Adaptive Averaging Pooling while maintaining its original resolution. Lastly, it performs the dimensional compression transformation by converting it to a 2D tensor following the flatten operation. The TCU retains vital information, excluding temporal fluctuations, and enables the network to overlook minor fluctuations in the temporal field. Consequently, it maintains the identification of facial contours characteristic in a reduced dimension.

\subsection{Cross-Task Feature Combiner (CTFC)}
\label{sec:cross-task feature combine block}
The Cross-Task Feature Combiner operates on the assumption that the pixel points representing physiological signals have minimal positional overlap with those of the facial profile used for identity recognition, and that the latter can surround the former in relative position.

As presented in Fig~\ref{fig:framework}, the CTFC takes in the avg-identity feature map $FM^{H_{i} \times W_{i}}_{i}$ and rPPG feature map $FM^{H_{r} \times W_{r}\times T_{r}}_{r}$ from the 2D and 3D Feature Extractors. Firstly, the resolution of the avg-identity feature maps is adjusted to be consistent through up-sampling.
The spatio-temporal dimensions of the identity feature map are then added and increased to match the shape of the rPPG feature map:
\begin{eqnarray}
FM^{H_{r} \times W_{r}\times T_{r}}_{i} = Cat(FM^{H_{r} \times W_{r}}_{i} ;...;FM^{H_{r} \times W_{r}}_{i})
\end{eqnarray}
Pixel-points are then summed after two 1x1x1 convolutions that maintain the overall pixel ratios of each feature map unchanged. The outputs of the resulting Fusion Feature Map $FuM$ are obtained.
\begin{eqnarray}
FuM^{H_{r} \times W_{r}\times T_{r}} &=& \alpha \bullet FM_{i} + \beta \bullet FM_{r}
\end{eqnarray}
with $\alpha$ and $\beta$ both scalars. The CTFC performs linear superposition of feature maps from different tasks by dimension adjustment and point-by-point convolution while preserving the original features. Fig~\ref{fig:gram-cam} shows that our CTFC allows the rPPG extractor to focus its attention inside the contour through task-independent contour information.

\subsection{Loss Function}
\label{sec:multi-task loss}
Based on~\cite{kendall2018multi}, a suitable multi-task loss function is employed for rFaceNet that maximizes the Gaussian likelihood with homologous uncertainty.
To balance the two tasks of extracting identity-specific facial contours and computing BVP signals based on fingertip extraction, heart rate derived from fingertip-extracted BVP is incorporated into the design loss. ECG signals are avoided to prevent potential conflicts with the BVP signal extraction task. The heart rate is calculated as the frequency corresponding to the maximum peak in the power spectrum obtained after performing a Fourier transform on the BVP signals.
\begin{eqnarray}
hr &=& argmax_{freq(0.5<, <4.2)} (|FFT(bvp)|^{2}) \times 60
\end{eqnarray}
In summary, it is assumed that the multiple outputs of the model consist of continuous outputs ${bvp}$, ${hr}$ and discrete outputs ${id}$, modelled by Gaussian likelihood and soft maximum likelihood, respectively. The ${Loss}(W, \sigma_{1}, \sigma_{2}, \sigma_{3})$ is as follows:
\begin{eqnarray}
\begin{aligned}
= -\log_{}{p(bvp,hr,id=gt_{id} |f^W(x))}\\
\approx \frac{1}{2\sigma_{1}^2} ||bvp -f^W(x)||^2+\log_{}{\sigma_{1}}\\
+\frac{1}{2\sigma_{1}^2} ||hr -f^W(x)||^2+\log_{}{\sigma_{2}}\\
+\frac{1}{\sigma_{3}^2}(-\log_{}{Softmax(id,f^W(x)))}+\log_{}{\sigma_{3}} 
\end{aligned}
\end{eqnarray}
And $\sigma_{1}$, $\sigma_{2}$ and $\sigma_{3}$ are positive scalars. The loss function aims to balance the training for both regression and classification tasks while preventing the premature learning of irrelevant information in the BVP signal prediction, which could otherwise result in significant fluctuations in the loss due to ambiguous identity information.

\section{Experiments}
\label{sec:experiment}

We first performed intra-dataset heart rate estimation experiment on the \textbf{VIPL-HR}~\cite{niu2019vipl} dataset. Then, to further demonstrate the validity of rFaceNet, we performed cross-dataset heart rate estimation experiments on the \textbf{UBFC-rPPG}~\cite{bobbia2019unsupervised} and \textbf{PURE}~\cite{stricker2014non} datasets using recently released public benchmarks~\cite{liu2022rppg}.

\subsection{Datasets and Evaluations}
The \textbf{VIPL-HR} database, designed for remote heart rate estimation from face videos, contains 2,378 VIS and 752 NIR videos from 107 subjects under various conditions. \textbf{UBFC-rPPG}, dedicated to rPPG analysis, includes 42 videos from 42 subjects without motion or lighting variations. \textbf{PURE} features recordings from 10 subjects (8 males, 2 females) across six movement conditions. 

The heart rate estimation is evaluated on \textbf{VIPL-HR} using Standard Deviation (SD), Mean Absolute Error (MAE), Root Mean Square Error (RMSE), and Pearson's Correlation ($\rho$). For \textbf{UBFC-rPPG} and \textbf{PURE}, the rPPG-toolbox~\cite{liu2022rppg} is used to compare performance with existing methods, employing MAE, RMSE, and $\rho$ as metrics.

\subsection{Implementation Details}
Our method, implemented in Pytorch, uses MTCNN on the \textbf{VIPL-HR} dataset to crop and stabilize the face region, excluding head motion scenes (\textbf{v2} and \textbf{v9}). The BVP and heart rate labels are upsampled to the original video sampling rate. During training, video frames are shaped into 64×128×128 segments, and following the approach suggested by~\cite{yu2022physformer}, the frames are divided into 3-4 segments of approximately 8 seconds each for testing, averaging the segment heart rates for the video-level heart rate. It is worth noting that we only normalize the RGB video frame data, without additional data enhancement, and train rFaceNet with a batch size of 16 using the Adam optimizer (initial learning rate 1e-5, weight decay 5e-5). 

To ensure fairness in benchmarking, the rppg-toolbox~\cite{liu2022rppg} is used for preprocessing, training, and evaluation in cross-dataset experiments. rFaceNet's input size is set to 128×72×72, and the resolution is doubled by interpolation from the TCU module. The experiments are conducted on the \textbf{UBFC-rPPG} and \textbf{PURE} datasets, using the same hyperparameters as PhysNet. Our method ultimately achieves state-of-the-art results on both benchmarks, highlighting the effectiveness of our approach.

\subsection{Intra-dataset Heart Rate Estimation}

\begin{table}[]
\centering
\footnotesize
\begin{tabular}{@{}ccccc@{}}
\toprule
\textbf{Method} & SD\(\downarrow\) & MAE\(\downarrow\) & RMSE\(\downarrow\) & $\rho$\(\uparrow\) \\ \midrule
Tulyakov2016\(\blacktriangle\)~\cite{tulyakov2016self} & 18.0 & 15.9 & 21.0 & 0.11 \\
POS\(\blacktriangle\)~\cite{wang2016algorithmic} & 15.3 & 11.5 & 17.2 & 0.30 \\
CHROM\(\blacktriangle\)~\cite{de2013robust} & 15.1 & 11.4 & 16.9 & 0.28 \\\midrule
RhythmNet\(\blacklozenge\)~\cite{niu2019rhythmnet} & 8.11 & 5.30 & 8.14 & 0.76 \\
ST-Attention\(\blacklozenge\)~\cite{niu2019robust} & 7.99 & 5.40 & 7.99 & 0.66 \\ 
NAS-HR\(\blacklozenge\)~\cite{lu2021hr} & 8.10 & 5.12 & 8.01 & 0.79 \\
CVD\(\blacklozenge\)~\cite{niu2020video} & 7.92 & 5.02 & 7.97 & 0.79 \\
Dual-GAN\(\blacklozenge\)~\cite{lu2021dual} & 7.63 & 4.93 & 7.68 & 0.81 \\ 
\midrule
I3D\(\ast\)~\cite{carreira2017quo} & 15.9 & 12.0 & 15.9 & 0.07 \\
PhysNet\(\ast\)~\cite{yu2019remote1} & 14.9 & 10.8 & 14.8 & 0.20 \\
DeepPhys\(\ast\)~\cite{chen2018deepphys} & 13.6 & 11.0 & 13.8 & 0.11 \\
AutoHR\(\ast\)~\cite{yu2020autohr} & 8.48 & 5.68 & 8.68 & 0.72 \\
PhysFormer\(\ast\)~\cite{yu2022physformer} & 7.74 & 4.97 & 7.79 & 0.78 \\
PhysFormer++\(\ast\)~\cite{yu2023physformer++} & 7.65 & \textbf{4.88} & 7.62 & 0.80 \\ 
rFaceNet\(\ast\) & \underline{\textbf{4.87}} &\underline{5.21} & \underline{\textbf{7.13}} & \underline{\textbf{0.87}}\\
\bottomrule
\end{tabular}
\centering
\caption{Intra-dataset Heart Rate estimation in VIPL-HR.The symbols \(\blacktriangle\), \(\blacklozenge\), and \(\ast\) denote the traditional approach, the non-end-to-end learning based approach, and the end-to-end learning based approach, respectively. The best results are marked in \textbf{bold} and our results are \underline{underlined}.}
\label{table:vipl}
\end{table}

\begin{figure}[t]
 \centering
 \footnotesize
 \includegraphics[page=3, width=0.4\textwidth, trim=3cm 3cm 3cm 3cm, clip]{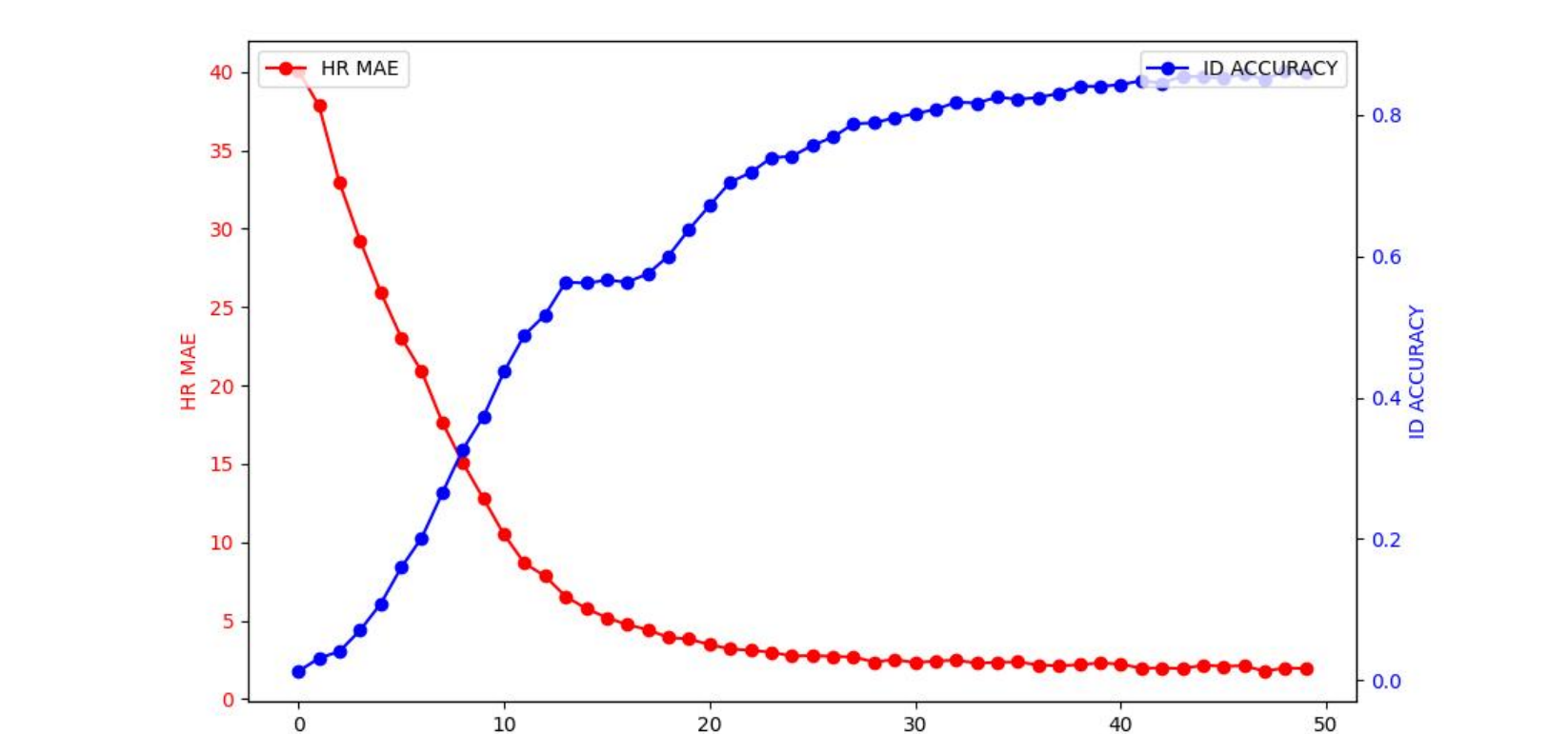}
 \caption{MAE of HR Estimation vs. Accuracy of identification.}
 \label{fig:train_loss}
 \vspace{-0.5cm}
\end{figure}
The table~\ref{table:vipl} illustrates that traditional methods often struggle with the VIPL-HR dataset due to complex environmental challenges. Deep learning models also face difficulties with weak physiological signals. However, our rFaceNet outperforms in heart rate estimation, leading in SD, RMSE, and $\rho$ metrics by effectively leveraging facial contour information. Despite being slightly behind in MAE, rFaceNet demonstrates robust performance in tracking heart rate trends under varying environmental conditions, proving its efficacy in complex scenarios.
The learning curves of rFaceNet is also shown in Fig~\ref{fig:train_loss}, visualizing the changes in MAE loss and facial contour-based identification accuracy for the heart rate estimation task during training. It is worth noting that the whole model first reaches an equilibrium point in the learning of the heart rate estimation task, which is then accompanied by a gradual convergence of the identification learning to the equilibrium and a synchronized fine-tuning of the heart rate estimation task.

\subsection{Cross-dataset Heart Rate Estimation}

\begin{table}[t]
\centering
\footnotesize
\begin{tabular}{@{}cccc@{}}
\toprule
\multicolumn{4}{c}{\textbf{Train: PURE \& Test: UBFC-rPPG}} \\ \midrule
\textbf{Method} & MAE\(\downarrow\) & RMSE\(\downarrow\)& $\rho$\(\uparrow\)\\ \midrule
GREEN\(\blacktriangle\)~\cite{verkruysse2008remote} & 19.73 ± 3.75 & 31.00 ± 235.38 & 0.37 ± 0.15\\
ICA\(\blacktriangle\)~\cite{poh2010advancements} & 16.00 ± 3.09 & 25.65 ± 163.58 & 0.44 ± 0.14 \\
CHROM\(\blacktriangle\)~\cite{de2013robust} & 4.06 ± 1.21 & 8.83 ± 33.93 & 0.89 ± 0.07 \\
LGI\(\blacktriangle\)~\cite{pilz2018local} & 15.80 ± 3.67 & 28.55 ± 236.17 & 0.36 ± 0.15 \\
PBV\(\blacktriangle\)~\cite{de2014improved} & 15.90 ± 3.25 & 26.40 ± 199.71 & 0.48 ± 0.14 \\
POS\(\blacktriangle\)~\cite{wang2016algorithmic} & 4.08 ± 1.01 & 7.72 ± 21.87 & 0.92 ± 0.06 \\ \midrule
TS-CAN\(\ast\)~\cite{liu2020multi} & 1.30 ± 0.40 & 2.87 ± 3.05 & \textbf{0.99 ± 0.02}\\
PHYSNET\(\ast\)~\cite{yu2019remote1} & 1.63 ± 0.53 & 3.79 ± 7.59 & 0.98 ± 0.03 \\ 
DEEPPHYS\(\ast\)~\cite{chen2018deepphys} & 1.21 ± 0.41 & 2.90 ± 3.75 & \textbf{0.99 ± 0.02}\\ 
EFF.PHYS-C\(\ast\)~\cite{liu2023efficientphys} & 2.07 ± 0.92 & 6.32 ± 32.01 & 0.94 ± 0.05 \\
rFaceNet\(\ast\) & \underline{\textbf{1.05 ± 0.35}} & \underline{\textbf{2.51 ± 2.55}} & \underline{\textbf{0.99 ± 0.02}} \\
\bottomrule
\end{tabular}
\centering
\label{table:intradataset}
\vspace{0.1cm}
\begin{tabular}{@{}cccc@{}}
\toprule
\multicolumn{4}{c}{\textbf{Train: UBFC-rPPG \& Test: PURE}} \\ 
\midrule
\textbf{Method} & MAE\(\downarrow\) & RMSE\(\downarrow\) & $\rho$\(\uparrow\) \\ \midrule
GREEN\(\blacktriangle\)~\cite{verkruysse2008remote} & 10.09 ± 2.81 & 23.85 ± 217.81 & 0.34 ± 0.12\\
ICA\(\blacktriangle\)~\cite{poh2010advancements} & 4.77 ± 2.08 & 16.07 ± 153.84 & 0.72 ± 0.09 \\
CHROM\(\blacktriangle\)~\cite{de2013robust} & 5.77 ± 1.79 & 14.93 ± 81.53 & 0.81 ± 0.08\\
LGI\(\blacktriangle\)~\cite{pilz2018local} & 4.61 ± 1.91 & 15.38 ± 134.14 & 0.77 ± 0.08 \\
PBV\(\blacktriangle\)~\cite{de2014improved} & 3.92 ± 1.61 & 12.99 ± 123.60 & 0.84 ± 0.07\\
POS\(\blacktriangle\)~\cite{wang2016algorithmic} & \textbf{3.67 ± 1.46} & \textbf{11.82 ± 66.87} & \textbf{0.88 ± 0.06} \\ \midrule
TS-CAN\(\ast\)~\cite{liu2020multi} & 3.69 ± 1.74 & 13.8 ± 113.84 & 0.82 ± 0.08\\
PHYSNET\(\ast\)~\cite{yu2019remote1} & 9.36 ± 2.39 & 20.63 ± 116.59 & 0.62 ± 0.10\\ 
DEEPPHYS\(\ast\)~\cite{chen2018deepphys} & 5.54 ± 2.30 & 18.51 ± 173.09 & 0.66 ± 0.10\\
EFF.PHYS-C\(\ast\)~\cite{liu2023efficientphys} & 5.47 ± 2.10 & 17.04 ± 143.80 & 0.71 ± 0.09\\
rFaceNet\(\ast\) & \underline{3.74 ± 1.52} & \underline{11.95 ± 71.94} & \underline{0.86 ± 0.07}\\
\bottomrule
\end{tabular}
\centering
\caption{Cross-dataset Heart Rate estimation in PURE and UBFC-rPPG.}
\label{table:crossdataset}
\vspace{-0.2cm}
\end{table}

\begin{figure}[t]
\centering
\footnotesize
 \includegraphics[page=4,width=0.52\textwidth, trim=1.5cm 3.5cm 0cm 3cm, clip]{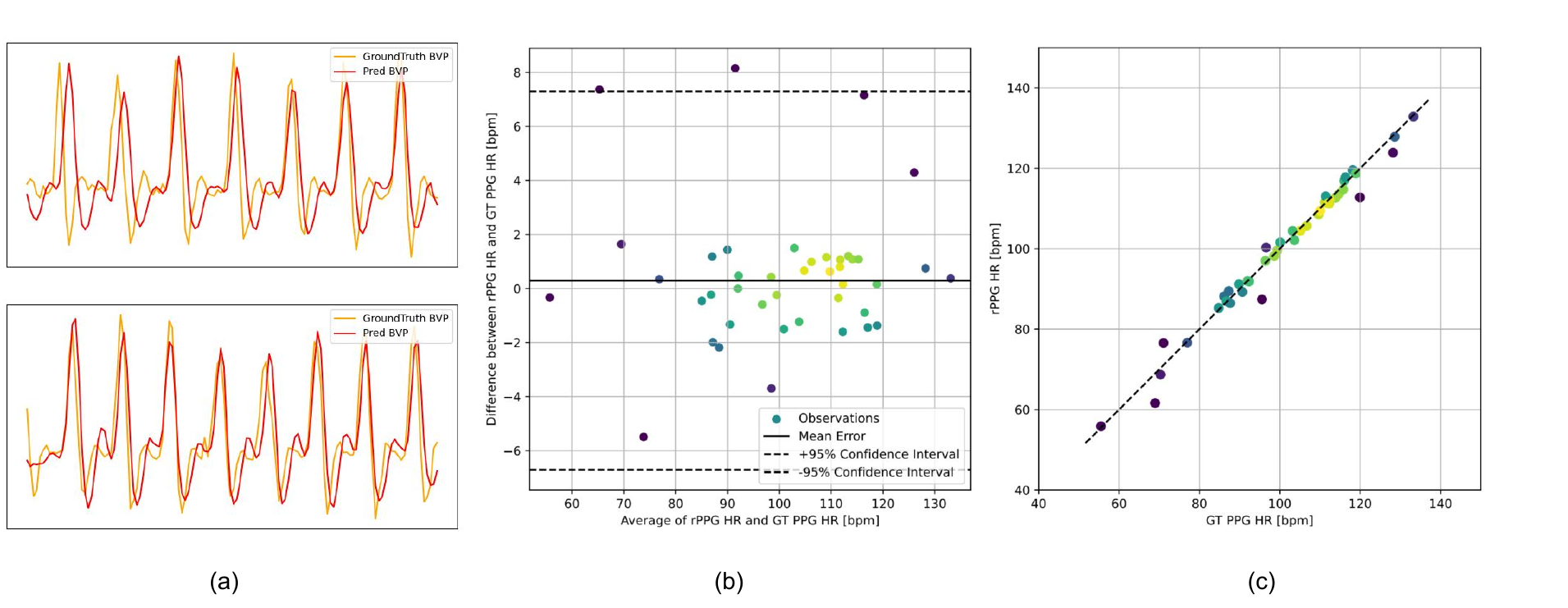}
 \caption{Predicted Signals vs. GroundTruth Signals.}
 \label{fig:comparion}
 \vspace{-0.5cm}
\end{figure}
As shown in Table~\ref{table:crossdataset}, rFaceNet performs excellently in cross-dataset heart rate estimation tests. Particularly, when using the \textbf{UBFC-rPPG} as the test set, it achieves state-of-the-art results in MAE, RMSE, and $\rho$ metrics. Similarly, when using \textbf{PURE} as the test set, rFaceNet maintains high accuracy, with only a slight difference from the optimal policy POS~\cite{wang2016algorithmic}, outperforming all deep learning schemes. This may be attributes to the fact that the \textbf{UBFC-rPPG} dataset has fewer scenes, and models trained on it struggle with the six different motion scene transformations in the \textbf{PURE} dataset.
Fig~\ref{fig:comparion}(a) shows that our extracted BVP signal closely matches the ground truth, especially in peak interval (heart rate) convergence, as reflected in the $\rho$ metric in Table~\ref{table:vipl}. Fig~\ref{fig:comparion}(b) presents a Bland-Altman plot comparing our method's heart rate estimation with true heart rate values. The plot indicates that the difference between the two methods stays within 8 bpm in the ±95\% confidence interval, demonstrating an acceptable level of agreement. Additionally, Fig~\ref{fig:comparion}(c) reveals a consistent pattern across a range of heart rate values, with most data points falling within the consistency range. This suggests that our method maintains accuracy across various heart rate values.

\subsection{Ablation Study}
\begin{figure}[t]
 \centering
 \footnotesize
 \vspace{-0.3cm}
 \includegraphics[page=5, width=0.5\textwidth, trim=1cm 6cm 1cm 5cm, clip]{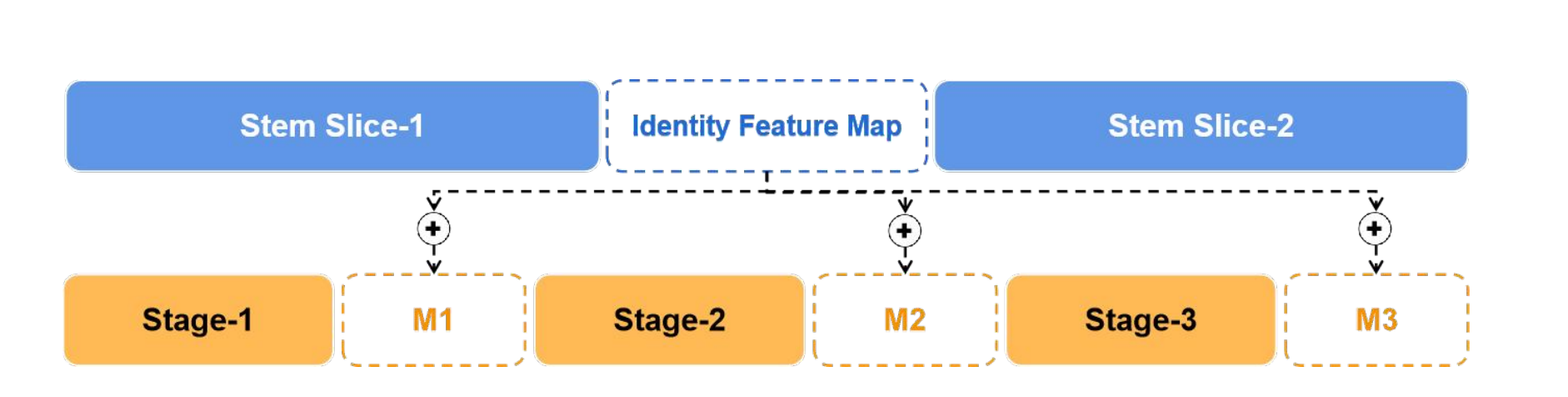}
 \caption{Three fusion schemes in CTFC design.}
 \label{fig:fusionschemes}
 \vspace{-0.2cm}
\end{figure}

\begin{table}[t]
\captionsetup{skip=5pt}
\centering
\footnotesize
\begin{tabular}{@{}ccccc@{}}
\toprule
\textbf{TCU}& \textbf{CTFC}& MAE\(\downarrow\) & RMSE\(\downarrow\) & $\rho$\(\uparrow\) \\ \midrule
\multirow{2}{*}{\centering \textbf{$\times$}} 
& \textbf{$\times$} & 1.99 ± 0.82 & 5.67 ± 24.84 & 0.95 ± 0.05 \\
& \textbf{\checkmark} & 1.57 ± 0.46 & 3.37 ± 4.17 & 0.98 ± 0.03 \\\midrule
\multirow{4}{*}{\centering \textbf{\checkmark}} 
& \textbf{$\times$} & 2.53 ± 0.88 & 6.26 ± 26.27 & 0.94 ± 0.05 \\
& \textbf{Stage-1} & 2.22 ± 0.91 & 6.30 ± 25.15 & 0.94 ± 0.06 \\
& \textbf{Stage-2} & 1.26 ± 0.43 & 3.04 ± 3.88 & 0.99 ± 0.03 \\
& \textbf{Stage-3} & \textbf{1.05 ± 0.35} & \textbf{2.51 ± 2.55} & \textbf{0.99 ± 0.02} \\
\bottomrule
\end{tabular}
\centering
\caption{Ablations of TCU and CTFC.}\label{table:tcu}
\vspace{-0.5cm}
\end{table}
In Table~\ref{table:tcu}, the TCU and CTFC parts of rFaceNet are split to highlight the advantages of combining the two. When only the CTFC part is added, random noise of the same shape replace the avg-identity feature map inputs, achieving improved results. This demonstrates that adding noise can enhance model stability. On the other hand, adding only the TCU module results in rFaceNet having two parallel task branches that share a common input but do not interact, leading to a performance drop in heart rate estimation due to the need to balance the training speeds of the two tasks. To experiment with the optimal design of the CTFC module, three fusion schemes are tested corresponding to the output feature maps M1, M2, and M3 from three stages of BVP signal extraction (stage 1, stage 2, and stage 3). As shown in Fig~\ref{fig:fusionschemes}, M1 has a resolution size of ${\left \lfloor{H^{input}/2},{W^{input}/2} \right \rfloor }$, M2 has a resolution size of ${\left \lfloor{H^{input}/4},{W^{input}/4} \right \rfloor }$, and M3 has a resolution size of ${\left \lfloor{H^{input}/8},{W^{input}/8} \right \rfloor }$.

According to Table~\ref{table:tcu}, the adding CTFC to the output M3 of Stage-3 yield optimal results. Fusion at the early stage of BVP signal extraction leads to decreased accuracy, possibly due to model instability during initial learning and increased noise from identity feature fusion at that stage. As demonstrated in Table~\ref{table:tcu}, our design proves superior, achieving optimal results when both components are combined.

\section{Conclusions}
\label{sec:conclusion}
In summary, our study introduces rFaceNet, a novel rPPG method that innovatively combines a Temporal Compressor Unit (TCU) to extract facial contours from temporally normalized frame inputs, thereby reducing redundant data. Additionally, we integrate a Cross-Task Feature Combiner (CTFC) to assimilate facial contour point information along the time dimension into physiological signal feature maps. This enhances the model's focus on pertinent facial regions, thereby improving the validity and interpretability of BVP signals extracted by rPPG. We demonstrate rFaceNet's superior performance in various benchmark tests. Furthermore, our study does not explore the potential integration of this mechanism with face recognition systems, which represents a promising direction for future research.


\bibliographystyle{IEEEbib}
\bibliography{main.bib}

\end{document}